\documentclass[runningheads]{llncs}

\usepackage{amssymb,amsmath,amsfonts}
\usepackage{graphicx}
\usepackage{url}
\usepackage{color}
\usepackage{amsfonts}
\usepackage{blindtext}
\usepackage{mathrsfs}
\usepackage{bbm}
\usepackage{url}
\usepackage{paralist}
\usepackage[english]{babel}
\usepackage{booktabs}
\usepackage{multicol}
\usepackage{multirow}
\usepackage{subfig}
\begin{document}
\begin{sloppypar}
\title{\textsc{SC-Ques}: A Sentence Completion Question Dataset for English as a Second Language Learners}

\author{
Qiongqiong Liu\inst{1}, 
Yaying Huang\inst{1}\thanks{Corresponding Author: Yaying Huang}, 
Zitao Liu\inst{2}, 
Shuyan Huang\inst{1}, 
Jiahao Chen \inst{1}, 
Xiangyu Zhao\inst{3}, 
Guimin Lin\inst{4}, 
Yuyu Zhou\inst{2}, 
Weiqi Luo\inst{2}}

\authorrunning{Q. Liu et al.}
%

\institute{TAL Education Group, Beijing, China \email{\{liuqiongqiong1,huangyaying1,huangshuyan,chenjiahao\}@tal.com} \and
Guangdong Institute of Smart Education, Jinan University, Guangzhou, China \email{\{liuzitao,zyy,lwq\}@jnu.edu.cn} \and
City University of Hong Kong, Hong Kong, China \\ \email{xianzhao@cityu.edu.hk} \and 
Shenzhen Everants Technology, Shenzhen, China \\ \email{lincank@everants.com} 
}

%
%

%
\maketitle              
\begin{abstract}
Sentence completion (SC) questions present a sentence with one or more blanks that need to be filled in, three to five possible words or phrases as options. SC questions are widely used for students learning English as a Second Language (ESL). In this paper, we present a large-scale SC dataset, \textsc{SC-Ques}, which is made up of 289,148 ESL SC questions from real-world standardized English examinations. Furthermore, we build a comprehensive benchmark of automatically solving the SC questions by training the large-scale pre-trained language models on the proposed \textsc{SC-Ques} dataset. We conduct detailed analysis of the baseline models performance, limitations and trade-offs. The data and our code are available for research purposes from: \url{https://github.com/ai4ed/SC-Ques}.
\end{abstract}

\section{Introduction}
\label{sec:intro}

Standardized examination is one of the crucial elements in worldwide education systems of teaching English as a Second Language (ESL) \cite{beinborn2015candidate,zweig2012computational}. They have proved a necessary source of evaluation data for investigating and diagnosing the situations that ESL learners grasp the essential language knowledge \cite{franke1960reform,madaus1991effects,davey2007university}. The standardized ESL examinations are usually organized in various formats of well-designed language processing tasks to evaluate specific capabilities. 

Previous researchers have spent lots of efforts in designing such language proficiency evaluation tasks which can be summarized into three categories: narrow, intermediate, and general \cite{zweig2012computational,woods2016exploiting,beinborn2016predicting}. These tasks are designed to assess different ESL language understanding capabilities from word, sentence, paragraph/document levels. For examples, among the narrow tasks, the identification of synonyms and antonyms has been widely used in the Test of English as a Foreign Language (TOEFL), Graduate Record Exams (GRE) and many other ESL standardized exams \cite{mohammad2008computing,turney2005corpus,turney2008uniform}. General tasks involve more logical and comprehensive abilities such as logic puzzles in the Law School Administration Test (LSAT) \cite{lev2004solving,lev2006logic,chesani2017solving}, reading comprehension questions from GRE \cite{ng2000machine,riloff2000rule}, etc. The intermediate tasks stand between the word-level narrow tasks and the paragraph-level general tasks and focus on the sentence level language understanding. Among all the intermediate tasks, sentence completion (SC) questions are well-studied as one of the classic representatives \cite{tang2016assessing,lee2014sentence,yang2019intelligent}. SC questions present a sentence with one or more blanks that need to be filled in. Three to five possible words (or short phrases) are given as options for each blank and only one of the options yields to a reasonable sentence. An example of SC question is shown in Table \ref{tab:example}.

\begin{table}[!bhpt]
\footnotesize
\vspace{-0.6cm}
\caption{An illustrative example of SC questions.} 
\begin{center}
\begin{tabular}{@{}ll@{}} \toprule
\label{tab:example}
 & — That T-shirt with Yao Ming's picture on it $\underline{\quad\quad}$ belong to John. \\
  & He likes  him a lot. \\ 
 & — No, it $\underline{\quad\quad}$ be his. He hates black color. \\
 & (A) can; can't (B) may; needn't (C) must; mustn't (D) must; can't \\  
\bottomrule
\vspace{-1.2cm}
\end{tabular}
\end{center}
\end{table}


An intelligent computational approach to SC in ESL is able to provide instant feedback to students and help students learn and practice ESL questions anytime anywhere. Besides, it provides feasible solutions to evaluate distractors in SC questions and helps teachers revise and improve the overall qualities of SC questions. 
Furthermore, the questions can also be used to generate more personalized questions that match the students' knowledge mastery, the mastery of students can be obtained by the knowledge tracing task \cite{liu2022pykt,liu2023simplekt,chen2023improving,liu2023enhancing}.
Hence, constructing a SC dataset is essential to building an intelligent SC approach.

Although there are a few publicly available SC datasets such as MSR SC dataset\footnote{https://www.microsoft.com/en-us/research/project/msr-sentence-completion-challenge/} \cite{zweig2011microsoft,tang2016assessing,woods2016exploiting}, their sample sizes are very limited and only contain a few thousand SC questions. Such small datasets are not able to align with the power of the state-of-the-art (SOTA) pre-trained language models (LMs). Furthermore, the number of missing blanks and the length of candidate tokens are fixed in existing open-sourced SC datasets \cite{zweig2011microsoft}. However, in the real-world English learning scenario, SC questions are usually presented in diverse forms with irregular numbers of blanks and various lengths of to-be-filled tokens.

To tackle the above limitations in existing SC datasets, we introduce \textsc{SC-Ques}, a large-scale SC question dataset for ESL learners. The proposed \textsc{SC-Ques} dataset contains 289,148 questions with one or more missing blanks. To the best of our knowledge, \textsc{SC-Ques} is one of the largest SC question dataset for ESL learners. Meanwhile, we fine-tune the existing SOTA pre-trained LMs on the \textsc{SC-Ques} dataset and present a systematic benchmark of automatically solving the ESL SC questions in students' real-life scenarios. We conduct comprehensive experimental evaluations to understand the performance impacts of different lengths of SC questions contexts and different numbers of words in candidate options. We conduct the precision-recall trade-off analysis and discuss the practical issues when deploying the SC question AI solver in real-world educational contexts.

\section{The SC-Ques Dataset}
\label{sec:dataset}
\subsection{Data Collection}

SC questions in \textsc{SC-Ques} are real English assignment questions used for K-12 students and they are developed by teaching content professionals from one of the largest educational technology companies in China. Each SC question in \textsc{SC-Ques} is represented as the single select multiple choice question format that consists of three parts: (1) question stem that denotes the main text content of the target ESL question with one or more blanks; (2) candidate options that are represented as a predetermined set of responses of at least three or more options; and (3) answer that indicates the correct one to be filled in the blanks.

Even though all the SC questions are manually developed by teaching professionals, there exists some data duplication, missing data problems or data corruption issues. Therefore, we conduct a series of steps to filter out or clean up the duplication and ill-formed SC questions. Specifically, we remove the duplicate question with the same options. And we filter out SC questions that if the question stem, candidate options, or answer is missing. Furthermore, we only remain SC questions whose number of candidate options is equal to 3 or 4.

\subsection{SC Question Categorization}

Due to the arbitrary number of blanks and tokens in the ESL SC tasks, questions in \textsc{SC-Ques} are categorized into the following categories according to the blank numbers and tokens numbers and the examples of each category are illustrated in Table \ref{tab:example2}. 

\begin{itemize}
\item \textbf{C1: One-blank and one-token.} Questions have one to-be-filled blank and the longest candidate option has one token.

\item \textbf{C2: One-blank and many-token.} Questions have one to-be-filled blank and the longest candidate option has multiple tokens.

\item \textbf{C3: Many-blank and one-token.} Questions have more than one to-be-filled blanks and the longest candidate option has one token.

\item \textbf{C4: Many-blank and many-token.} Questions have more than one to-be-filled blanks and the longest candidate option has multiple tokens.
\end{itemize}

\vspace{-0.55cm}

\begin{table}[!hbpt]
\footnotesize
\vspace{-0.3cm}
\caption{Illustrative examples of SC questions from different categories.}
\vspace{-0.3cm}
\begin{center}
\scalebox{0.9}{
\begin{tabular}{c|l}\toprule
\label{tab:example2}
\multirow{4}{*}{  C1  } &  Jack is five years old now. He can dress $\underline{\quad\quad}$. \\
~ & (A) herself (B) himself (C) yourself \\
\cline{2-2}
~ & Sam and Mike aren't at school. $\underline{\quad\quad}$ are they? \\
~ & (A) How (B) Who (C) Where \\
\hline
\hline
\multirow{6}{*}{  C2  } & — I'm very tall. Where can I sit? \\
~ & — $\underline{\quad\quad}$ the classroom. \\
~ & (A) In the front of (B) At the back of (C) In the middle of \\
\cline{2-2}
~ & — I find that I have been unlucky these days. \\
~ & — $\underline{\quad\quad}$ Everything will become  better soon. \\
~ & (A) Keep silent! (B) Why not cry? (C) Cheer up! (D) How about you? \\
\hline
\hline
\multirow{7}{*}{  C3  } & — Whose are those jeans? \\ 
~ & — This pair of jeans $\underline{\quad\quad}$ mine and that two pairs $\underline{\quad\quad}$ my brother's. \\
~ & (A) are; are (B) are; is (C) is; are  \\ 
\cline{2-2}
~ & — That T-shirt with Yao Ming's picture on it $\underline{\quad\quad}$ belong to John. \\
~ & He likes  him a lot. \\ 
~ & — No, it $\underline{\quad\quad}$ be his. He hates black color. \\
~ & (A) can; can't (B) may; needn't (C) must; mustn't (D) must; can't \\
\hline
\hline
\multirow{8}{*}{  C4  } & We have $\underline{\quad\quad}$ homework to do every day so we can't play $\underline{\quad\quad}$ \\ 
~ & computer games. \\ 
~ & (A) too many; too many (B) too much; too many \\ 
~ & (C) too many; too much (D) too much; too much \\ 
\cline{2-2}
~ & $\underline{\quad\quad}$ of their bad habits, the boys changed from being dependent on their \\
~ & parents to $\underline{\quad\quad}$ for others. \\
~ & (A) Having rid; sacrificing (B) To rid; sacrifice \\
~ & (C) Rid; sacrificing (D) Having been rid; sacrifice \\
\bottomrule
\end{tabular}
}
\vspace{-0.6cm}
\end{center}
\end{table}

\subsection{Data Statistics}

\begin{figure*}[!htbp]
\footnotesize
\centering
\includegraphics[width=0.85\textwidth]{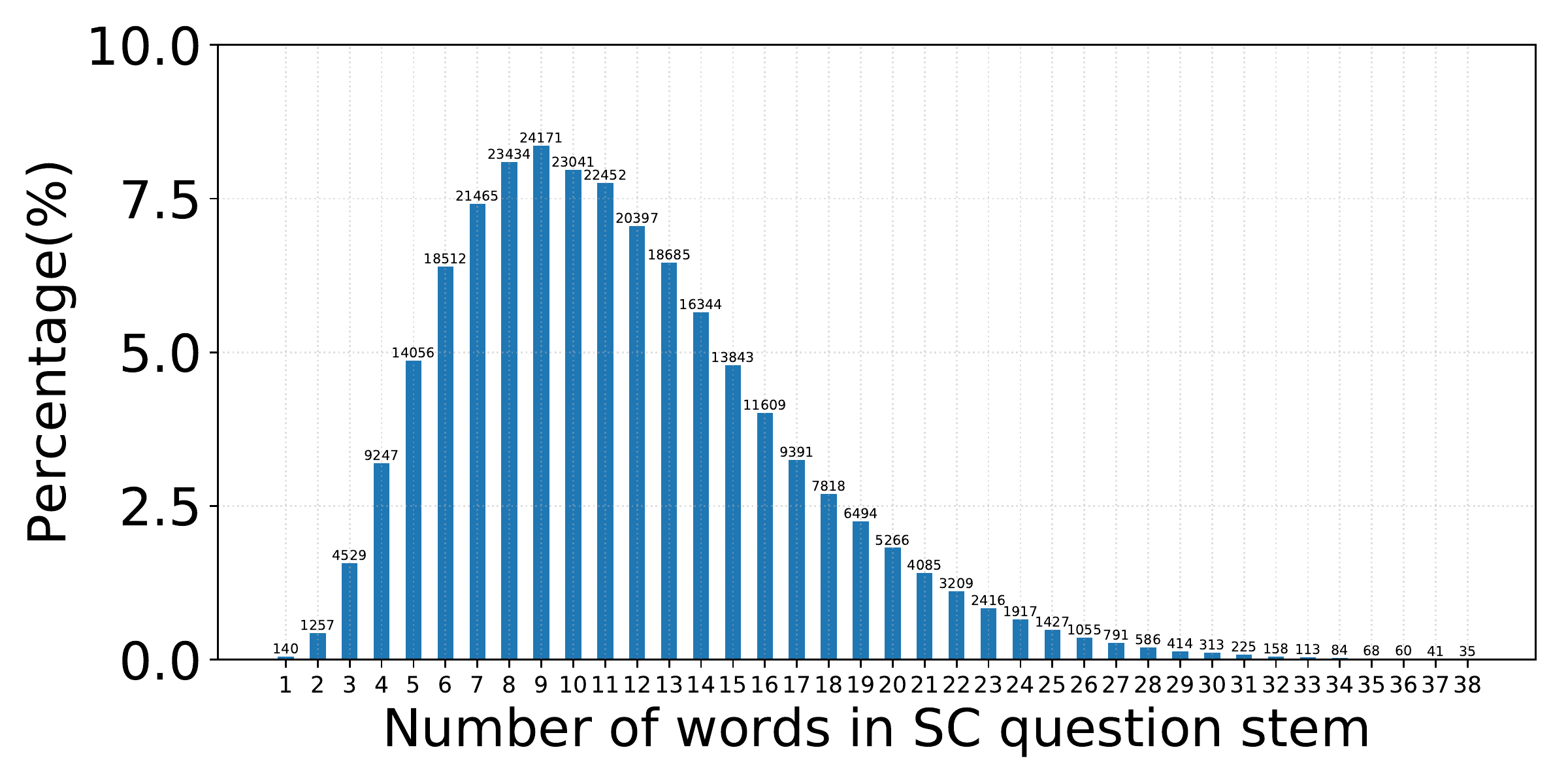} 
\caption{Length distribution of SC questions in \textsc{SC-Ques}.}\label{fig:len_distribution}
\end{figure*}

After data cleaning and preprocessing, in total, we end up with 289,148 SC questions in \textsc{SC-Ques}. Specifically, we have 110,645 133,249, 27,886, and 17,368 SC questions in categories C1, C2, C3 and C4 respectively. 84.35\% of SC questions have one blank missing. The length distribution of the question stems in \textsc{SC-Ques} is shown in Figure \ref{fig:len_distribution}. As we can see that most of the questions contain 5 to 15 words and sentences of 9 words have the largest portion. There are only 5.88\% questions that have more than 20 words. This indicates SC questions usually have very limited contexts and imply subtle linguistic knowledge such as grammar, syntax, and semantics, the performance of automatically solving the ESL SC questions may vary a lot.


\section{Benchmark}
\label{sec:method}

\subsection{Problem Formulation}
\label{sec:problem}

Let $\mathbf{q}$ be the SC question stem with one or more blanks. Let $\mathbf{o}_1, \cdots, \mathbf{o}_m$ be the candidate options associated with $\mathbf{q}$. Solving the SC question is to find the option that leads to the highest correct probability after completing the to-be-filled sentence with the selected option, i.e., $\arg\max_{i = 1, \cdots, m} \mbox{Pr}(\mathbf{o}_i | \mathbf{q})$. 

\subsection{Benchmark Workflow}

\begin{figure*}[!hbpt]
\centering
\includegraphics[width=0.98\textwidth]{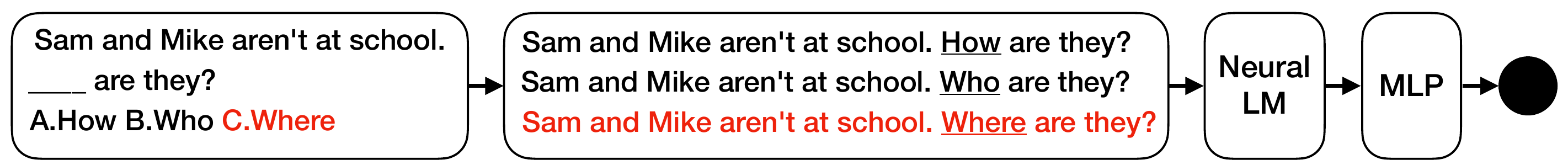}
\caption{The overall benchmark workflow. The sentence in red color denotes the correct option and is denoted as the positive example in the training process.}\label{fig:train}
\end{figure*}

In this work, we view the above problem as a predictive task and we aim to train a classifier to find the correct answer from the option pool of confusing candidates. Specifically, we first fill candidate options into the corresponding blanks to get complete sentences. Then we treat sentences that contain the correct options as positive examples and the rest as negative examples. After that, we utilize a neural LM model to extract the semantically meaningful information within each sentence in SC questions and make final SC question predictions via a multilayer perceptron (MLP). The overall end-to-end benchmark workflow is shown in Figure \ref{fig:train}.


\noindent \textbf{The Neural Language Model}. Large-scale pre-trained LMs such as BERT \cite{kenton2019bert}, RoBERTa \cite{liu2019roberta}, XLNet \cite{yang2019xlnet}, BART \cite{lewis2020bart} and DeBERTa \cite{he2021deberta}, benefit from self-supervised learning on a large amount of corpus, and has shown its competent generalization to various downstream natural language understanding tasks. In this work, we conduct a systematic benchmark evaluation by measuring the prediction accuracy with existing prevalent large-scale pre-trained LMs listed as follows. Please note that we choose to use their large model versions for experiments. Since we expect each SC question to be solved correctly, we choose to use accuracy as our evaluation metric. 

\begin{itemize}
\item BERT \cite{kenton2019bert}. A pre-trained natural language understanding model with transformer encoder blocks. We fine-tune BERT on our SC dataset as a sentence classification task. We use a special start token ([CLS]) as the first token of every text sequence and the final hidden state corresponding to this token is used as the aggregated SC sentence representation.
\item RoBERTa \cite{liu2019roberta}. RoBERTa improves BERT by replacing static masking with dynamic masking, pre-training more epochs with larger batch size, and removing the next sentence prediction task. We follow the same fine-tuning protocol as described in the BERT method.
\item XLNet \cite{yang2019xlnet}. XLNet is an autoregressive based pre-training method with transformer decoder blocks. Similar to BERT, we fine-tune the XLNet and utilize the last hidden state as the SC representation.
\item BART \cite{lewis2020bart}. BART adapts standard Transformer \cite{vaswani2017attention} as its backbone model and utilizes a denoising autoencoder for pretraining sequence-to-sequence models. It is pre-trained to recover different types of text corruptions to their original versions, such as sentence permutation and text infilling.
\item DeBERTaV3 \cite{he2023debertav}. DeBERTaV3 improves the original DeBERTa \cite{he2021deberta} model by replacing mask language modeling with replaced token detection. The model is trained as a discriminator to predict whether a token in the corrupted input is either original or replaced by a generator.
\end{itemize}

\noindent \textbf{The MLP Prediction}. Once we obtain the final hidden state $\mathbf{t}_n$ from above neural LM module as the aggregated SC sentence representation. We introduce two additional fully-connected layers to perform the binary classification task, i.e., $\mathbf{x} = \mathrm{softmax}(\mathbf{W}_1 \mathrm{tanh}(\mathbf{W}_0 \mathbf{t}_n + \mathbf{b}_0) + \mathbf{b}_1))$, where $\mathbf{W}_0$, $\mathbf{W}_1$, $\mathbf{b}_0$ and $\mathbf{b}_1$ are trainable parameters, $\mathbf{W}_0 \in \mathbb{R}^{1024 \times d}$, $\mathbf{b}_0 \in \mathbb{R}^ {1024}$, $\mathbf{W}_1 \in \mathbb{R}^{2 \times 1024}$ and $\mathbf{b}_1 \in \mathbb{R}^{2}$, $\mathbf{t}_n \in \mathbb{R}^{d}$. The first entry of $\mathbf{x}$ gives the probability of wrong option while the second entry gives right option probability. The objective is to minimize the cross entropy of the right or wrong option labels.

\subsection{Experimental Setup \& Details}

We randomly split the entire dataset into training set and testing set with 241,195 and 47,953 SC questions respectively in an approximate ratio of 5 : 1. For BERT, RoBERTa, XLNet and DeBERTaV3, each model has 24 hidden layers, 1024 hidden size and 16 attention heads. For the BART model, it has 12 hidden layers and 16 attention heads in each of the encoder and decoder, and the hidden size is 1024. We employ the AdamW \cite{kingma2014adam} as the optimizer with an initial learning rate of 1e-5 for all models. We fine-tune our model on 4 Tesla V100 GPU devices. Due to the limited memory of GPU devices, the batch size of BERT is set to 32, and the other models set the batch size as 16. The max length of the sentences is set to 128. 

\subsection{Results}

\begin{figure*}[!htbp]
\footnotesize
\centering
\includegraphics[width=0.85\textwidth]{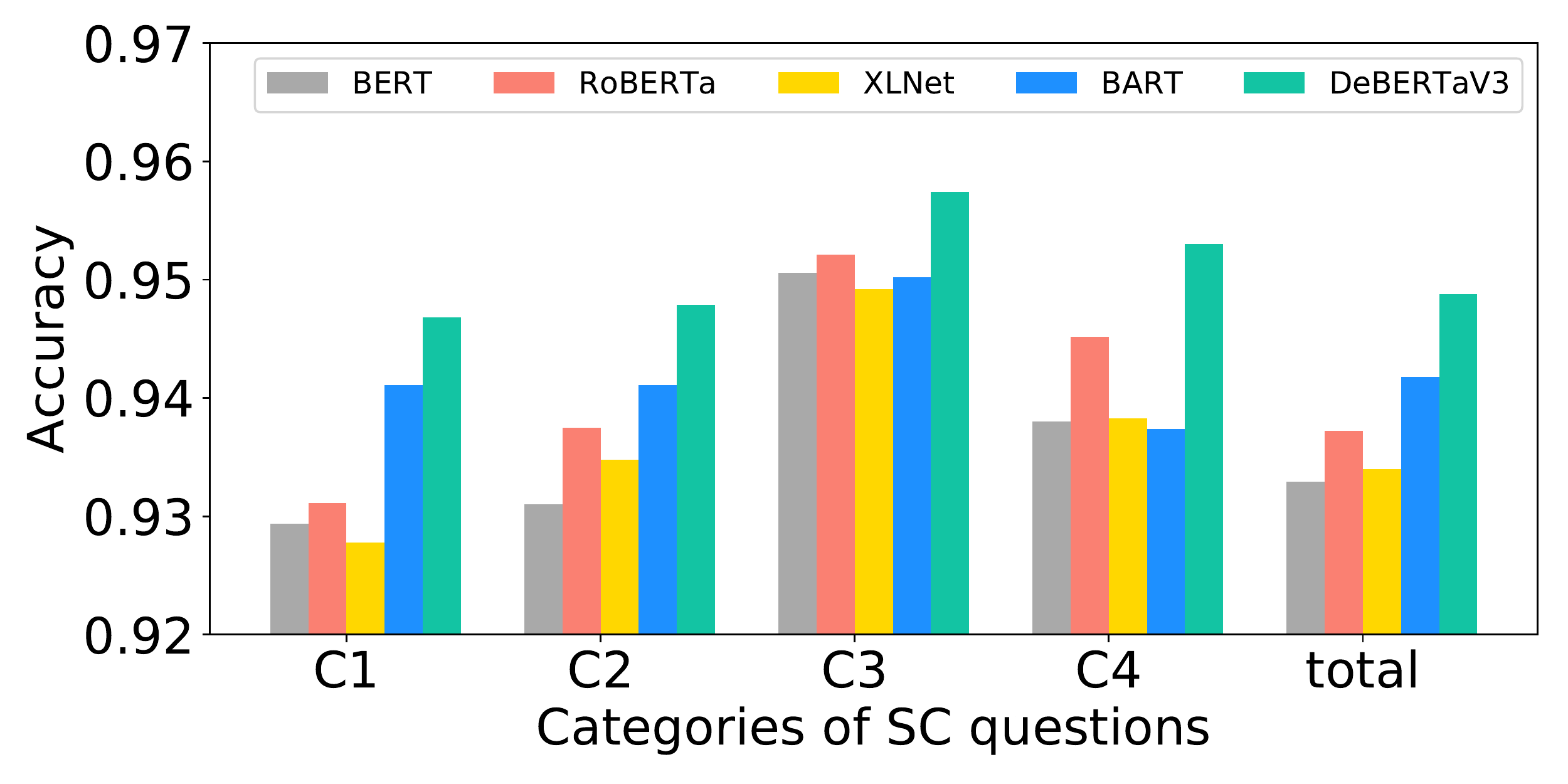} 
\caption{Accuracy on different categories of \textsc{SC-Ques}.}\label{fig:overall_res}
\end{figure*}

\noindent \textbf{Overall Performance}. As we can see from Figure \ref{fig:overall_res}, DeBERTaV3 outperforms other models in terms of prediction errors on the entire testing set. This indicates DeBERTaV3 has better generalization performance among all the selected pre-trained LMs which may be due to the effective token detection task to discriminate whether the corrupted token is original or replaced by a generator in DeBERTaV3. 
When comparing the prediction performance of all the methods on C1, C2, C3 and C4, we can see that C3 and C4 have better performance than C1 and C2, we think the reason is that C1 and C2 have only one blank, the difference of the completed sentences of each option are quite similar which makes the models hard to distinguish this subtle difference.
We run the pairwise t-test for each possible pair from these methods and all of them are statistically significantly different at 0.01 level. 


\vspace{-0.5cm}

\vspace{-0.8cm}

\begin{figure*}[!htbp]
\footnotesize
\centering
\includegraphics[width=0.85\textwidth]{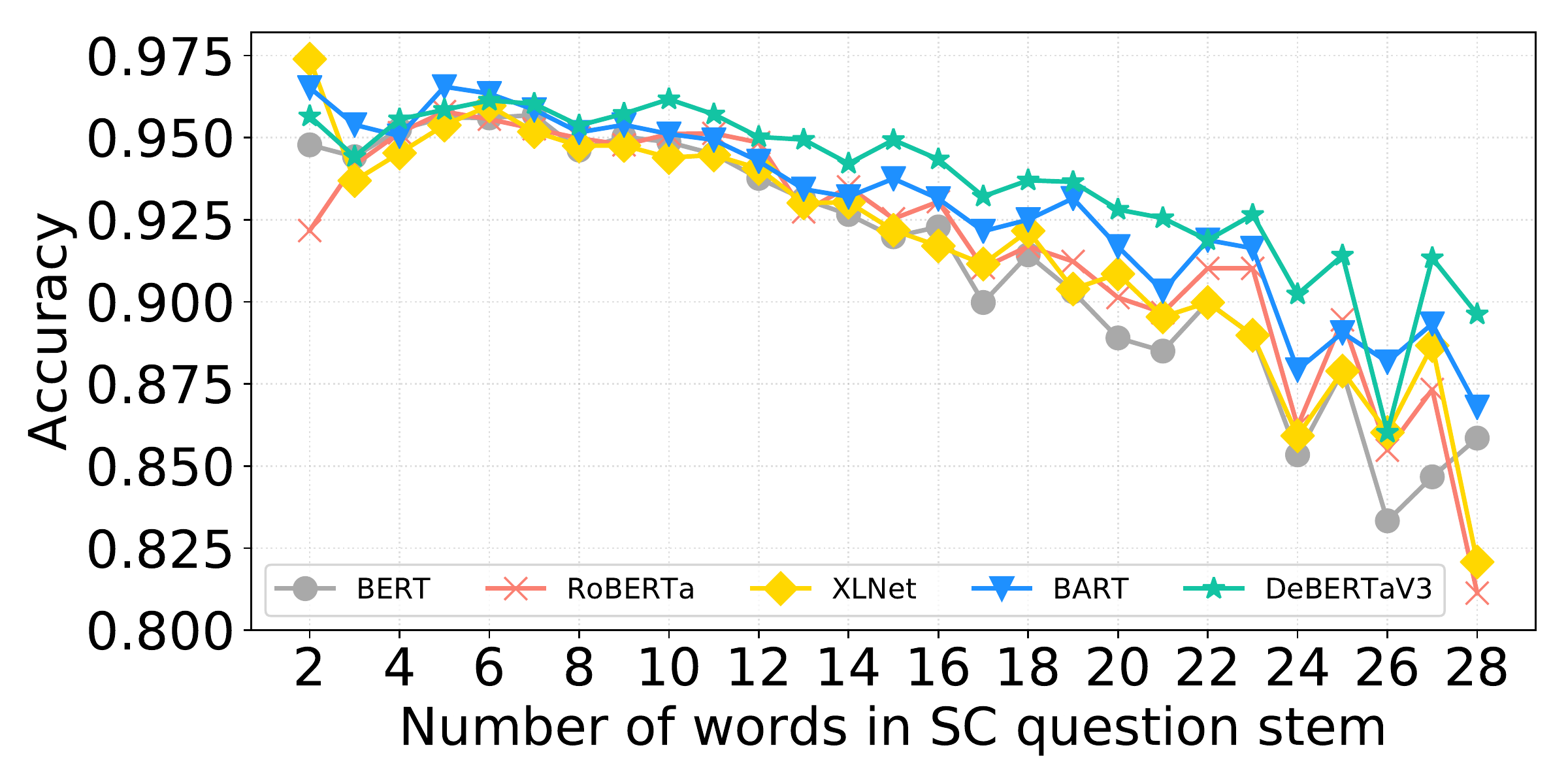} 
\caption{Length distribution of SC questions in \textsc{SC-Ques}.}\label{fig:acc}
\end{figure*}


\noindent \textbf{Impacts of Different Lengths of SC Questions Contexts}. We study the impacts of different lengths of questions and the results are shown in Figure \ref{fig:acc}. From the model performance curve, when the length of the questions is larger than 10, the models' performances drop sharply. This indicates that after exceeding a certain length, questions become more difficult with the increase of the context length. The reason may be that the model is difficult to capture the contextual information in such a long distance. What is interesting is that if the length of the question is less than 3, the models also show slightly poor performance, which indicates that it is hard for the model to distinguish different options once there is not enough context information. More specifically, the performance of DeBERTaV3 is less than the performance of BART when the length of the question is less than 7. With the increasing sequence length, DeBERTaV3 outperforms other models. We believe the reason is that DeBERTaV3 is trained as a discriminator to detect whether the token is original, the longer sequences provide relatively effective information to make DeBERTaV3 discriminate the tokens, so they can get higher prediction results of SC compared to other models.


\noindent \textbf{Impacts of Different Numbers of Words In Candidate Options}. We also study the performance impacts in terms of the length of candidate options. We focus on the questions which have only one blank and two blanks. As shown in Table \ref{tab:BlankResults}, we can see that the trend is similar to the different lengths of the questions, DeBERTaV3 achieves the best performance in all lengths of candidate options. If the length of the candidate options is too short or too long, the performance would be a little worse.

\vspace{-0.3cm}

\begin{table}[!bhpt]
\footnotesize
\caption{Accuracy on different lengths of the longest option in the questions which have \emph{one blank} and \emph{two blanks}.} 
\label{tab:BlankResults}
\vspace{-0.4cm}
\begin{center}
\setlength{\tabcolsep}{1mm}{ 
\scalebox{0.9}{
\begin{tabular}{lccccccccc}
\toprule
                             & \multicolumn{4}{c}{one blank}                                          &  & \multicolumn{4}{c}{two blanks}                                         \\ \cline{2-5} \cline{7-10} 
\multirow{-2}{*}{Methods}    & 1               & 2               & 3               & \textgreater{}=4 &  & 2               & 3               & 4               & \textgreater{}=5 \\ \hline
BERT                         & 0.9294          & 0.9433          & 0.9307          & 0.8984           &  & 0.9496          & 0.9388          & 0.9489          & 0.9161           \\
RoBERTa                      & 0.9311          & 0.9481          & 0.9345          & 0.9130           &  & 0.9527          & 0.9467          & 0.9506          & 0.9367           \\
XLNet                        & 0.9278          & 0.9439          & 0.9323          & 0.9136           &  & 0.9483          & 0.9324          & 0.9489          & 0.9288           \\
BART                         & 0.9411          & 0.9505          & 0.9393          & 0.9184           &  & 0.9503          & 0.9388          & 0.9465          & 0.9272           \\
DeBERTaV3                    & \textbf{0.9468} & \textbf{0.9543} & \textbf{0.9450} & \textbf{0.9349}  &  & \textbf{0.9568} & \textbf{0.9571} & \textbf{0.9571} & \textbf{0.9383}  \\
\bottomrule
\end{tabular}}
}
\end{center}
\end{table}

\vspace{-0.3cm}

\begin{figure*}[htbp]
\footnotesize
\centering
\includegraphics[width=0.85\textwidth]{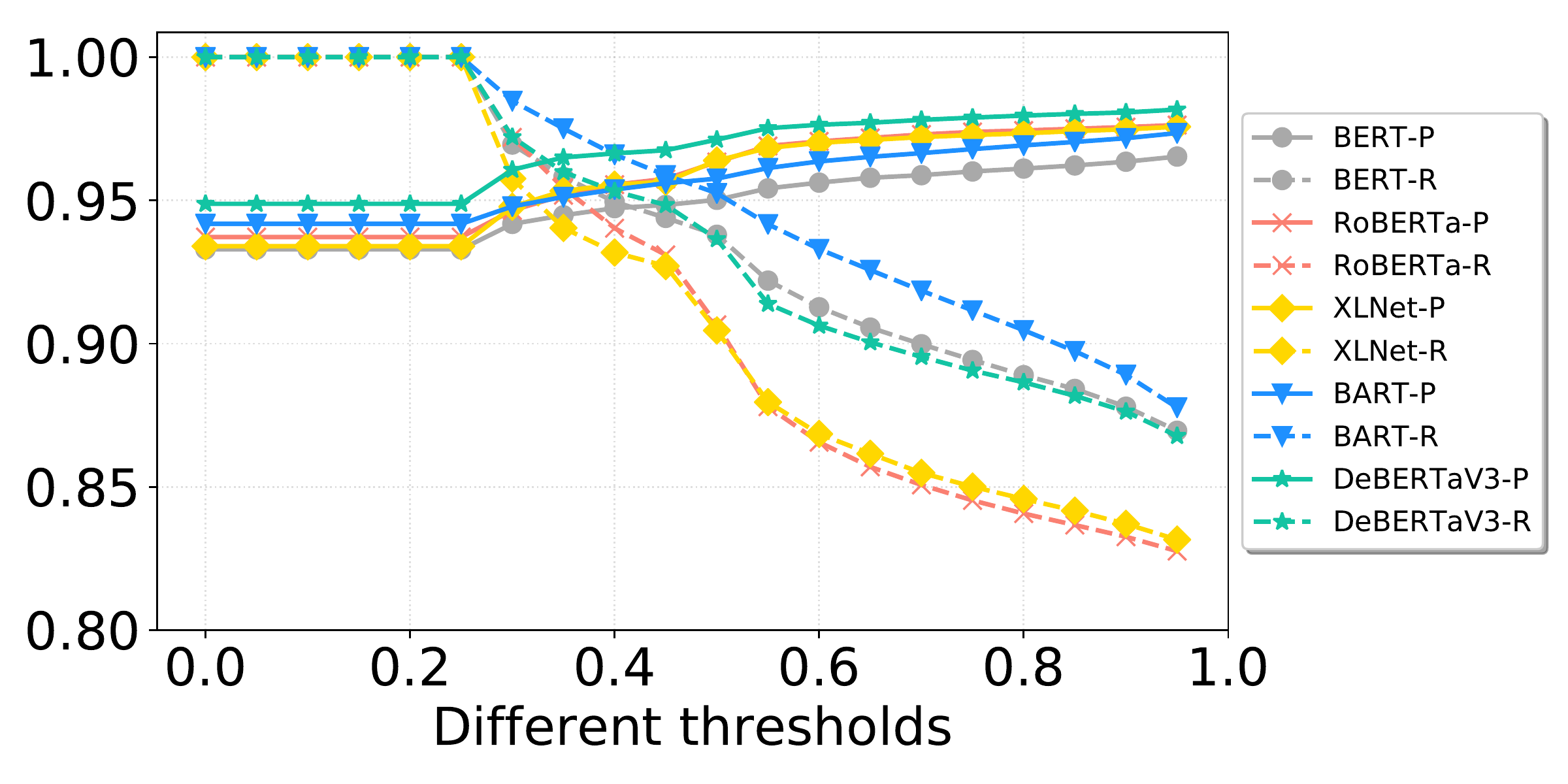}
\caption{The precision-recall curve.}\label{fig:pr}
\vspace{-0.8cm}
\end{figure*}

\noindent \textbf{Precision-Recall Trade-off Analysis}. When deploying the model in practice, a wrong answer may give bad guidance to students. In order to reduce such problems, we may refuse to solve some difficult questions and improve the precision of more solvable questions. The models in our benchmark will output the correct probability for each option, for each question, we use a softmax function with temperature 0.1 for the probabilities of all options, the final highest probability can also be viewed as the problem-solvable confidence score. After that, we set a threshold to the correct probability of the model's selected option and accept the above-the-threshold questions as our solvable questions. The recall score is computed as (the number of solvable questions)/(the number of all test questions), and the precision score is calculated as (the number of both solvable and correct-answered questions)/(the number of solvable questions). The precision and recall curves of all the pre-trained LMs in different thresholds are shown in Figure~\ref{fig:pr}, We can see that with the threshold growing, the precision is higher and the recall becomes smaller. When the threshold is 0.95, the precision of most of the models are higher than 97.0\% and the recall scores keep greater than 80.0\%. Especially, regardless of the threshold values, the precision of DeBERTaV3 is always higher than 95.0\% and the recall scores never lower than 87.0\%.

\section{Related Work}
\label{sec:related} 
\subsection{Sentence Completion Datasets}

SC datasets always include in-completed sentences which require selecting the most appropriate answer from several options. Recently, there are several publicly available SC datasets. MSR is a SC challenge data consisting of 1,040 sentences with four impostor sentences for each original sentence \cite{zweig2011microsoft}. CODAH involves multiple choice SC questions, each SC question contains four options which are consistent with commonsense \cite{chen2019codah}. Park and Park collected 1823 cloze-style questions with multiple-choice questions from the Test of Proficiency in Korean (TOPIK) \cite{park2020assessment}. Nozza et al. provided a manually labeled hurtful SC dataset in six languages \cite{nozza2021honest}. Compared with the above works, our proposed \textsc{SC-Ques} is a relatively large-scale SC dataset with more than 290 thousand ESL SC questions.


\subsection{Sentence Completion Solvers}


Various approaches have been proposed to automatically solve the ESL SC questions. Earlier literature prefer complete the sentence by filling the blank with each candidate option in turn, and to evaluate its likelihood under an LM \cite{zweig2012computational,chen2009performance,chen2009shrinking}. Motivated by the successful applications in language modeling via deep neural networks \cite{schuster1997bidirectional,graves2013speech,mikolov2011extensions}, recent studies tackle the SC challenges based on the recurrent neural networks (RNN) \cite{tran2016recurrent,yang2019intelligent}. For examples, Yang and Deng proposed a global context dependent RNN LM to obtain the global semantics of the target in the SC tasks \cite{yang2019intelligent}. Since large-scale pre-trained LMs become remarkably powerful models for language understanding and attain unprecedented levels of performance gains on multiple language tasks \cite{Peters2018DeepCW,howard2018universal,kenton2019bert,radford2019language}, there are some approaches utilize pre-trained LMs to solve the SC questions \cite{shen2020blank,donahue2020enabling,zhu2019text}. Donahue et al. trained the language model by using the concatenation of artificially-masked texts and the texts which are masked as input \cite{donahue2020enabling}. Besides above research works of solving SC questions via language modeling, researchers also make attempts from different machine learning perspectives \cite{woods2016exploiting,argouarc2018dependency,gubbins2013dependency,fedus2018maskgan,Banerjee2013MultipleCQ,liu2019tigs}. Liu et al. developed an gradient search based iterative inference algorithm that can be broadly applied to any neural sequence generative model for the fill-in-the-blank tasks \cite{liu2019tigs}. Researchers also studied automatically solve questions in domains other than ESL, such as history \cite{wang2014cmu}, science \cite{khot2018scitail}, etc.

\section{Conclusion}
\label{sec:conclusion}
In this paper, we construct a large-scale SC question dataset, \textsc{SC-Ques}, for ESL Learners. The \textsc{SC-Ques} consists of 289,148 SC questions with four categories of questions that contain almost all SC classifications in the English examinations of the real world. Furthermore, we present a benchmark on the proposed \textsc{SC-Ques} for automatically solving the ESL SC questions. The experimental results indicate the effectiveness of our dataset and the proposed benchmark.

\section*{Acknowledgments}
This work was supported in part by National Key R\&D Program of China, under Grant No. 2020AAA0104500; in part by Beijing Nova Program (Z201100006820068) from Beijing Municipal Science \& Technology Commission and in part by NFSC under Grant No. 61877029 and in part by Key Laboratory of Smart Education of Guangdong Higher Education Institutes, Jinan University (2022LSYS003).
%
%
%
%
\bibliographystyle{splncs04}
\bibliography{its2023}

\end{sloppypar}
\end{document}